%% file: 00-main.tex
\pdfoutput=1

\documentclass[11pt]{article}

\usepackage[]{acl}

\usepackage{times}
\usepackage{latexsym}
\usepackage{booktabs}
\usepackage{pdfpages}

\usepackage[T1]{fontenc}

\usepackage[utf8]{inputenc}

\usepackage{microtype}
\usepackage{amsmath}
\usepackage{amssymb}
\usepackage{bm}
\usepackage{todonotes}
\usepackage{graphicx}
\makeatletter
\def\thickhline{%
  \noalign{\ifnum0=`}\fi\hrule \@height \thickarrayrulewidth \futurelet
   \reserved@a\@xthickhline}
\def\@xthickhline{\ifx\reserved@a\thickhline
               \vskip\doublerulesep
               \vskip-\thickarrayrulewidth
             \fi
      \ifnum0=`{\fi}}
\makeatother
\newlength{\thickarrayrulewidth}
\usepackage{comment}
\usepackage{multirow}
\usepackage{subcaption}
\usepackage{tabularx}
\usepackage{xcolor}
\usepackage{pdfpages}

\newcommand{\corpus}{\texttt{AB\reflectbox{B}A}\,}

\title{On the Potential of Debiasing Adapters for Computational Argumentation}
\title{Investigating Debiasing Adapters for Computational Argumentation}
\title{Bias Evaluation and Mitigation for Computational Argumentation}
\title{Argumentative and Fair Language Modeling for\\ Computational Argumentation}
\title{Fair and Argumentative Language Modeling \\for Computational Argumentation}

\author{Carolin Holtermann\textsuperscript{1}, Anne Lauscher\textsuperscript{2}, Simone Paolo Ponzetto\textsuperscript{1} \\
  \textsuperscript{1}Data and Web Science Group, University of Mannheim, Germany \\
  \textsuperscript{2}MilaNLP, Bocconi University, Italy \\
  \texttt{cholterm@mail.uni-mannheim.de} \\
  \texttt{anne.lauscher@unibocconi.it} \\
  \texttt{simone@informatik.uni-mannheim.de} }

\date{}

\begin{document}
\maketitle
\begin{abstract}
Although much work in NLP has focused on measuring and mitigating stereotypical bias in semantic spaces, research addressing bias in computational argumentation is still in its infancy. In this paper, we address this research gap and conduct a thorough investigation of bias in argumentative language models. To this end, we introduce \corpus, a novel resource for bias measurement specifically tailored to argumentation. We employ our resource to assess the effect of argumentative fine-tuning and debiasing on the intrinsic bias found in transformer-based language models using a lightweight adapter-based approach that is more sustainable and parameter-efficient than full fine-tuning. Finally, we analyze the potential impact of language model debiasing on the performance in argument quality prediction, a downstream task of computational argumentation. Our results show that we are able to successfully and sustainably remove bias in general and argumentative language models while preserving (and sometimes improving) model performance in downstream tasks. We make all experimental code and data available at \url{https://github.com/umanlp/FairArgumentativeLM}.
\end{abstract}

\section{Introduction}
\input{01-intro}

\section{\corpus: A New Annotated Corpus of Bias in Argumentative Text}
\input{02-data}

\section{Adapter-based Fair Argumentative Language Models}
\input{03-methodology}

\section{Experiments and Results}
\input{04-experiments}

\section{Related Work}
\input{06-rw}

\section{Conclusion}
\input{07-conclusion}

\section*{Acknowledgments}
The work of Anne Lauscher is funded by the  European Research Council (ERC) under the European Union’s Horizon 2020 research and innovation program (grant agreement No. 949944, INTEGRATOR). We thank the anonymous reviewers for their insightful comments.

\section*{Limitations and Further Ethical Considerations}
\input{08-limitations}

\bibliographystyle{acl_natbib}
\bibliography{references}

\clearpage
\newpage
\section*{Supplementary Material}
\appendix
\input{xx-appendix}

\end{document}

%% file: 01-intro.tex
\label{sec:intro}

Recently, pre-trained language models (PLMs), e.g., BERT~\citep{BERT}, RoBERTa \citep{roberta}, GPT-2~\citep{GPT2} and DialoGPT~\citep{DialoGPT} have been shown to encode and amplify a range of stereotypical biases, such as racism, and sexism~\citep[e.g.,][\emph{inter alia}]{kurita-etal-2019-measuring, dev2020measuring, nangia-etal-2020-crows, lauscher2021sustainable}. While such types of biases provide the basis for interesting academic research, e.g., historical analyses~\citep[e.g.,][\emph{inter alia}]{garg2018word, tripodi2019tracing, walter2021diachronic}, stereotyping constitutes a representational harm~\citep{barocas2017problem,blodgett-etal-2020-language}, and can lead in many concrete socio-technical application scenarios to severe ethical issues by reinforcing societal biases  \citep{hovy-spruit-2016-social, shah-etal-2020-predictive, Mehrabi21}.

But while prior work has focused on how to evaluate and mitigate unfair biases for general-purpose LMs~\citep[e.g.,][]{webster2020measuring} and their applications to specific domains and genre like, for instance, conversational LMs~\citep[e.g.,][]{barikeri2021redditbias}, there has been little attention to the problem of \textit{bias in argumentative language}. This is despite previous work from \citet{spliethoever} pointing out the high potential for harm, due to the high sensitivity of envisioned applications like self-determined opinion formation systems, as well as, crucially, showing that argumentative corpora like those from the online debate portal \texttt{debate.org}~\citep{debateOrgcorpus} do encode unfair biases, which are likely to be captured by argumentative LMs. This is particularly problematic as research in computational argumentation regularly makes use of such corpora for injecting knowledge about argumentative language into PLMs~\citep[e.g.,][]{alshomary-etal-2021-belief}. Still, to date, there is neither an evaluation resource specifically tailored to argumentative language, nor knowledge on debiasing argumentative LMs or on the effects of debiasing on argumentative downstream tasks.

\paragraph{Contributions.} We address this research gap with the following contributions: we present \corpus, the first human-annotated resource specifically targeted at English argumentative language, which is annotated for two kinds of social bias that are still under-explored in NLP, namely \emph{Queerphobia} and \emph{Islamophobia}. Next, we use  \corpus to answer the following four research questions (\textbf{RQs}):

\vspace{1.2mm}
\noindent \textbf{(RQ1)} \textit{How does argumentative fine-tuning affect measurable biases in PLMs?}

\vspace{1.2mm}
\noindent We show that the impact of argumentative fine-tuning can induce and increase measurable stereotypical biases in the LMs, highlighting the importance of bias measurement after injecting argumentative knowledge (\S\ref{sec:rq1}).

\vspace{1.2mm}
\noindent \textbf{(RQ2)} \textit{Can we validate the effectiveness and efficiency of debiasing PLMs using adapters?}

\vspace{1.2mm}
\noindent \citet{lauscher2021sustainable} recently introduced \emph{debiasing adapters}, a modular and sustainable way of encoding debiasing knowledge in LMs. We confirm the effectiveness of debiasing adapters with Counterfactual Data Augmentation~\citep{zhao_etal_2018_cda} on two diverse corpora (\S\ref{sec:rq2}).

\vspace{1.2mm}
\noindent \textbf{(RQ3)} \textit{Can we obtain an (efficient and robust) fair and argumentative language model given our preexisting set of adapters?}

\vspace{1.2mm}
\noindent We show for the first time how to stack debiasing adapters with argumentation adapters to produce an \textbf{argumentative and fair language model}. Our results indicate that stacking order matters (\S\ref{sec:rq3}).

\vspace{1.2mm}
\noindent \textbf{(RQ4)} \textit{What are the effects on argumentative downstream tasks, e.g., argument quality prediction?}

\vspace{1.2mm}
\noindent In a final downstream evaluation encompassing two different datasets for argument quality prediction, we demonstrate that debiasing can have a positive impact on model performance. On one of the corpora, our best results are obtained when combining argumentation and debiasing adapters, hinting at the effectiveness of fair and argumentative language modeling (\S\ref{sec:rq4}).

We hope that our results and our novel \corpus resource will fuel more research on fair computational argumentation.

%% file: 02-data.tex
\label{sec:data}

We create \corpus, the first annotated corpus of bias in argumentative text following the methodology from \citet{barikeri2021redditbias}: (1) specification of the social biases of interest, (2) retrieval of candidates of biased statements, and (3) manual annotation.

\paragraph{Bias Specifications.} \label{sec:BiasSpecification}
We define the social biases we are interested in using the established notion of explicit bias specifications \citep{Caliskan_2017, DEBIE}. It consists of two sets of target terms $(T_1$ and $T_2)$ denoting two demographic groups that exhibit different stereotypical perceptions w.r.t. two opposing sets of attribute terms $(A_1$ and $A_2)$. Concretely, $T_1$ consists of target terms referring to a minoritized group (e.g., \textit{Muslim}), while $T_2$ consists of target terms corresponding to a dominant group (e.g., \textit{Christian}), i.e., a group in power~\citep{d2020data}. %
We focus on the bias dimensions \textit{Queerphobia} and \textit{Islamophobia} since they have received little attention in NLP research on bias when compared to sexism or other ethnic bias.
We view \textit{Queerness} as an umbrella term for the minority group of the \textit{LGBTQI+} community, which includes people of all sexual orientations and gender identities except for heterosexual and cisgender. We compare this to the dominant group of heterosexual cisgender people.

\setlength{\tabcolsep}{3pt}
\begin{table*}[t]
    \centering
    \small
    \begin{tabular}{lllll}
        \toprule
          \textbf{Dimension} & \multicolumn{2}{c}{\textbf{Target Term Sets}} & \multicolumn{2}{c}{\textbf{Attribute Term Sets}}\\
        \midrule
        \multirow{2}{*}{Islamophobia} & \textbf{\bm{$T_1$}} & \emph{muslim(s)}, \emph{islam}, \emph{quran}, \emph{koran}, ...   &  \textbf{\bm{$A_1$}} & \emph{terrorist}, \emph{rapist}, \emph{enemy}, \emph{bomb}, \emph{oppressed}, ... \\
        & \textbf{\bm{$T_2$}} & \emph{christian(s)}, \emph{christianity}, \emph{bible}, \emph{church}, ...   & \textbf{\bm{$A_2$}}  &\emph{police}, \emph{friend}, \emph{defend}, \emph{peace}, \emph{safety}, ...  \\
        \midrule
        \multirow{2}{*}{Queerphobia} & \textbf{\bm{$T_1$}} & \emph{gay(s)}, \emph{lesbian(s)},  \emph{queer(s)}, \emph{bisexual(s)}, ... &\textbf{\bm{$A_1$}}  & \emph{weak}, \emph{immoral}, \emph{fashion}, \emph{sinful}, ...  \\ 
        & \textbf{\bm{$T_2$}} & \emph{straight(s)}, \emph{hetero(s)}, \emph{heterosexual(s)} \emph{cisgender(s)}, ...  & \textbf{\bm{$A_2$}} & \emph{strong}, \emph{moral},  \emph{scientific}, \emph{healthy}, ... \\
        \bottomrule
    \end{tabular}
    \caption{\texttt{AB\reflectbox{B}A} bias specifications for candidate retrieval.}
    \label{tab:biasTerms}
\end{table*}

The target and attribute terms used for candidate identification are based on the specifications of \citet{barikeri2021redditbias}. They include a wide range of attribute terms from the sociological literature and manually compiled target terms. The attribute terms were assembled such that each stereotypical attribute term $a_1$ forms a loose antonym of an counter-stereotypical attribute term $a_2$ with a positive or negative sentiment.
An exemplary partial term list of the bias specifications can be found in Table \ref{tab:biasTerms} and the full set in the Appendix.

\paragraph{Candidate Retrieval.} We use the dataset from \texttt{debate.org} originally collected by \citet{debateOrgcorpus}, one of most widely used resources in research on computational argumentation.

For retrieving candidates, we compute the Cartesian product of the terms of the minoritized group $T_1$ with all stereotypical terms of $A_1$, giving us a set of stereotyped tuples from $T_1 \times A_1$ (e.g., \emph{gay} and \emph{sinful}). Using this set, we extract all sentences and their corresponding arguments that contain both terms from the tuples in a window of size 20 (set during corpus construction to improve the quality of the retrieved passages). We further reduced the compiled comments to those with a maximum number of 500 tokens to allow for a better visualization and to ensure that the annotators attentively read the entire argument. In total, we retrieve 889 candidate sentences from 614 different arguments for \emph{Queerphobia} and 1,879 candidate sentences from 1,101 different arguments for \emph{Islamophobia}.

\paragraph{Annotating bias.} We manually label the candidate sentence and the corresponding argument according to whether a stereotypical bias is present or not. %
To this end, we hired four annotators, who are all non-native speakers but have excellent English proficiency with academic backgrounds and who hold at least a Bachelor's degree, in slightly different majors (engineering, data science, information systems, and computer science). They are of diverse gender and cultural background. %

Annotators were provided with the guidelines found in the Appendix. We initially conducted a pilot study on 90 randomly drawn arguments to iteratively calibrate annotations and refine the guidelines on the basis of the annotators' feedback. Finally, we split the corpus evenly into four independent, equally-sized portions and added further $50$ randomly drawn overlapping arguments to analyze annotation quality. In the last step, we merged the annotations on the calibration set using majority voting. The number of annotated and biased instances in the corpus is shown in Table \ref{tab:ABBA}. We show examples of biased sentences in Table~\ref{tab:ex_biased}.

\setlength{\tabcolsep}{2.5pt}
\begin{table}[t]
    \centering
    \small{
    \begin{tabular}{lcccc}
        \toprule
          & \multicolumn{2}{c}{\textbf{Sentence-level}} & \multicolumn{2}{c}{\textbf{Argument-level}}\\
        \cmidrule(lr{0.2em}){2-3} \cmidrule(lr{0.2em}){4-5} 
        \textbf{Dimension} & \# ann. & \# bias. & \# ann. & \# bias. \\
        \midrule
        \textbf{Islamophobia}  & 1,860 & 648 (34.84\%)  & 1,090 & 333 (30.55\%)\\
        \textbf{Queerphobia} & 862 &  358 (41.65\%) & 601 & 205 (34.11\%)\\
        \bottomrule
    \end{tabular}}
    \caption{Total number of annotated (\# ann.) and biased (\# bias.) sentences and arguments in \corpus.}
    \label{tab:ABBA}
\end{table}

\begin{table*}[t]
\centering
\small{
\begin{tabularx}{\textwidth}{lXl}
\toprule
    Dimension   &  Example Sentence & Label\\
\midrule
\multirow{1}{*}{\textbf{Islamophobia}} & \textit{6 billion \underline{muslims} around the world are following the religion of \underline{violence}, \underline{hate} and \underline{terror.}} & Biased\\
& \textit{I would agree that there should be punishments for \underline{terrorism}, but not for \underline{Islam} itself.} & Unbiased\\
 \multirow{1}{*}{\textbf{Queerphobia}} & \textit{Thus, since being \underline{gay} is a \underline{sin} and sins are poor choices, being \underline{gay} is a choice.} & Biased\\
 & \textit{The stigma of \underline{homosexuals} being more \underline{promiscuous} is a horrible lie.} & Unbiased\\
 \bottomrule
\end{tabularx}}
\caption{Example sentences from \corpus.}\label{tab:ex_biased}
\end{table*}

\paragraph{Analysis of the Annotations.} On the overlapping set consisting of $50$ arguments, we obtain an inter-annotator agreement (IAA) for \emph{Queerphobia} on the sentence-level for both Fleiss' $\kappa$~\citep{FleissKappa} and Krippendorff's $\alpha$~\citep{Krippendorff} of $0.65$. The agreement on the argument-level is slightly weaker with $0.61$ for both measures. For the \textit{Islamophobia} dimension, we observe a stronger agreement of $0.66$ on sentence-level and $\kappa = 0.72$ and $\alpha = 0.73$ on the argument-level. Although we are dealing with a rather subjective annotation task, IAA indicates a \textit{substantial} agreement among the annotators \citep{InterobserverAgreement}, suggesting that they are able to reliably identify stereotypes in argumentative sentences and longer text. %

To determine reasons for disagreement among annotators, we manually conducted a qualitative analysis on the annotated arguments. For \textit{Queerphobia}, we found that annotators mostly disagreed on statements that referred to the homosexual lifestyle, rather than homosexual people. The following example illustrates one such case: %

\begin{quote}
    \textit{[...] Basically, a gay person is not allowed to engage in sexual acts with another man because there is a 0\% chance of offspring being produced. This falls into the same category of not using contraceptives, getting abortions, etc. \textbf{It is not a sin for a gay person to acknowledge their sexuality, or to act in a `gay' manner.} It is only a sin if he/she gives in to their urges. [...] }
 \end{quote}

 Here, the annotators disagreed in the annotation of the entire argument. Although the debater clearly states that actually being gay is not a sin, in his opinion, living a homosexual lifestyle is a sin. It appears that for some annotators being homosexual is equivalent to living in a homosexual relationship, while others clearly distinguished these two aspects. %
For \textit{Islamophobia}, the disagreements mostly related to arguments that make a distinction between Muslims and the religion Islam, e.g.:

\begin{quote}
    \textit{\textbf{[...] I have no issue with Islam, or any religion in general, if you leave me alone I leave you alone, you wondered why so many people hate Islam, its because of the same [...] in your last paragraph, y'all act as if terrorism is 100\% okay.} That needs to change before Muslims can consider Islam anywhere close to a great religion.}
\end{quote}

Here, the fact that the debater is making an ambiguous statement, expressing no prejudice against Islam but against Muslims caused confusion among the annotators resulting in disagreement. %

%% file: 03-methodology.tex
\label{sec:method}

To obtain a fair and argumentative LM, we conduct both argumentative and debiasing language modeling along our two bias dimensions of interest. Instead of full model fine-tuning, we opt for a more sustainable strategy by relying on adapters \citep{Adapters}
to reduce computation time and energy consumption. In addition, the modularity of adapters enables their reuse in further settings and in combination with other pre-trained adapters.

\paragraph{Argumentation Adapter.} \label{sec:CondLM}

Following \citet{alshomary-etal-2021-belief}, we tune general pre-trained models on a large set of arguments to obtain an argumentative language model. In contrast to the original work, we rely on language adapters. Concretely, we adopt the architecture proposed by \citet{MADX}, which inserts a single adapter, a two-layer feed-forward network, into each transformer layer. The output of the adapter is computed as %
\begin{equation*}%
    \textnormal{A}_\textnormal{{argument}}(\mathbf{h},\mathbf{r})= \mathbf{U}(\text{ReLU}(\mathbf{D}(\mathbf{h}))) + \mathbf{r}\,,
\end{equation*}
with the two matrices $\mathbf{D} \in \mathbb{R}^{h \times d}$ and $\mathbf{U} \in \mathbb{R}^{d \times h}$ as the adapter's down-projection and up-projection, respectively, $\mathbf{h}$ as the transformer's hidden state, and $\mathbf{r}$ as the residual. In addition, we inject invertible adapters, which are stacked on top of the embedding layer and the inverses of the invertible adapters are placed in front of the output layer. They perform a similar function to the language adapters, but aim to capture token-level specific transformations \citep{MADX}. Both the language adapters and the invertible adapters are trained on a language modeling task using a causal language modeling loss for auto-regressive models and a masked language modeling loss for auto-encoding models, respectively.

\paragraph{Debiasing Adapter.} \label{sec:debiasingAdapter}
For debiasing, we inject debiasing adapters~\citep{lauscher2021sustainable} into the models, using the same adapter architecture as before. Following the original work, we use Counterfactual Data Augmentation~\citep[CDA]{zhao_etal_2018_cda} and train the adapter parameters on the augmented corpus to break stereotypical associations in the model. 

To this end, we manually compile pairs of opposing target terms $(t_i,t_j) \in T_1 \times T_2$, such that $t_i$ forms the most suitable antonym of $t_j$ in the sense of minority and dominant group (e.g., \emph{muslim} and 
\emph{christian}) and can be substituted grammatically interchangeably. While this is arguably straightforward with the \textit{Islamophobia} bias specifications, the target terms of the \textit{Queerness} dimension are more complex to juxtapose. Therefore, we clustered them into three groups of `sexual identity' (e.g., \textit{\{gay, straight\}}), `gender identity' (e.g., \textit{\{transgender, cisgender\}}) and `biological sex' (e.g., \textit{\{androgyne, unisexual\}}) so as to find the best matching pairs of antonyms (cf.\ the list in the Appendix). %
We then replace all occurring target terms from $T_1$ or $T_2$ with their opposite term from the set of tuples $P = \{(t_i,t_j)\}^N$ (we randomly select a term from the list if multiple substitutions are possible).

We opt for a two-sided application of CDA, keeping both the counterfactual and the original sentences in the training set to avoid over-correction \citep{webster2020measuring}. We append each counterfactual sentence immediately after its original counterpart and train in two settings, namely using: a) only biased and counterfactual sentences; b) all sentences, i.e.,  also including neutral ones.

\paragraph{Combining Adapters.} \label{sec:debiasingMethod}
\label{sec:combining}
We investigate three different architectures: first, in \S \ref{sec:rq3}, we study two architectures using \textit{AdapterStacking}~\citep{MADX}, i.e., by stacking the argumentation adapter on top of a debiasing adapter and \emph{vice versa} (Figure~\ref{fig:adapterStacking}). Second, in \S \ref{sec:rq4}, we compare the best architectures from \S \ref{sec:rq3} with \textit{AdapterFusion}~\citep{MADX}, which requires training additional network layers for interpolating the adapters' outputs.

%% file: 04-experiments.tex
\label{sec:experiments}

We next describe the experiments to answer the research questions RQ1 through RQ4 (Section \ref{sec:intro}) that underpin our investigation.

\subsection{Measuring the Effect of Argumentative Fine-tuning}
\label{sec:rq1}

\paragraph{Language Model Bias (LMB) Score.} We follow \citet{barikeri2021redditbias} and employ \corpus for computing the LMB score reflecting how much more likely the model is to generate a stereotypically biased argument compared to an inversely biased one. We start with our set of opposing target terms $P \subset T_1 \times T_2$ and we extract the set of all statements $S$ from \corpus (containing instances of term $t_i$ such that $(t_i, t_j) \in P$), which have been labelled as stereotypically biased. 
This results in $279$ biased instances for  \textit{Queerphobia} and $465$ instances for \textit{Islamophobia}, respectively. We then create for each instance $s_{(t_i, a)} \in S$ (e.g., \textit{All Muslims are terrorists}), a corresponding inversely biased sentence $s'_{(t_j, a)}$ (e.g., \textit{All Christians are terrorists}) to give us a set $S'$ of counter-stereotypical statements. %
In case of multiple pairs for a target term (e.g., \textit{\{homosexual, heterosexual\}} and \textit{\{homosexual, straight\}}), we create one counter-stereotypically biased sentence for each possible combination. We then compute the model's perplexity for all statements in the two paired sets $S$ and $S'$ with stereotypical and counter-stereotypical statements.  Following \citet{barikeri2021redditbias}, we compute the mean perplexity for multiple counterfactual instances created from a single biased instance and remove outliers to avoid distorted significance results \citep{outlier}. The final LMB score corresponds to the t-value obtained by subjecting the paired perplexities to the student's t-test ($\alpha = 0.05$).

\begin{figure}[t]
    \centering
    \includegraphics[width=5.5cm]{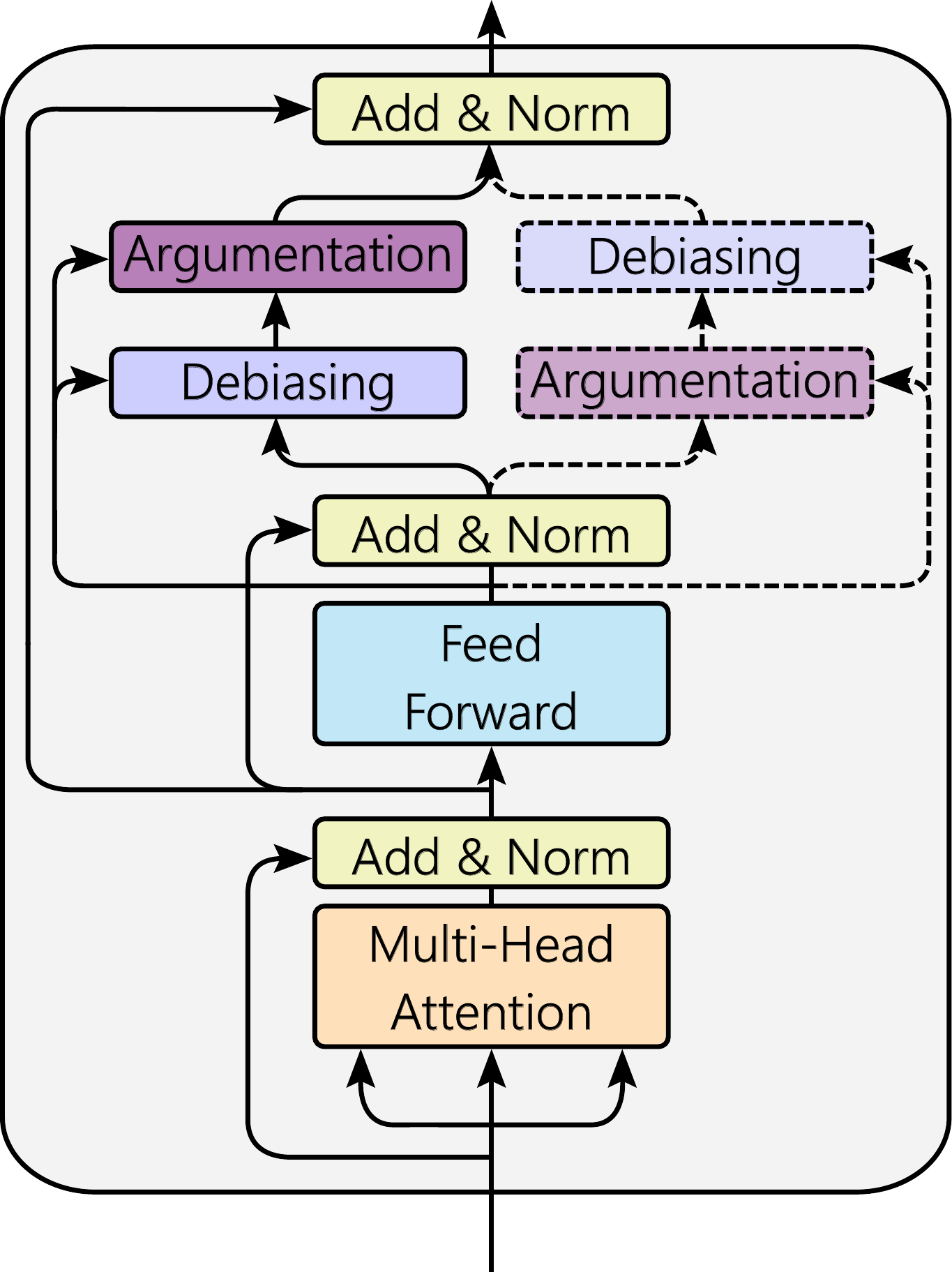}
    \caption[Adapter Stacking Architecture]{AdapterStacking architectures.}
    \label{fig:adapterStacking}
\end{figure}

\paragraph{Fine-tuning Data.} We test the effect of argumentative fine-tuning using two argumentative corpora: (i) Args.me \citep{argsme}, which consists of over 380k arguments from over 59k debates. (ii) Considering that it contains mostly arguments retrieved from Debate.org ($\sim 87 \%$), we verify our results using a second corpus: Webis-ChangeMyView-20~\citep[CMV;][]{CMV}, which contains over 3.6 million arguments extracted from the ChangeMyView subreddit. %
For ensuring comparability, we cut each corpus to 300k and perform a train-validation split of 80:20.

\paragraph{Models.} We experiment with four LMs from Huggingface Transformers~\citep{wolf-etal-2020-transformers}:
BERT ({\small\texttt{bert-base-uncased}}), %
GPT-2 ({\small\texttt{gpt-2}}), %
DialoGPT ({\small\texttt{microsoft/DialoGPT-medium}}) and %
RoBERTa ({\small\texttt{roberta-base}}). %
With the exception of DialoGPT, which contains contains $24$ layers with a hidden size of $1,024$, all  models consist of $12$ layers with a hidden size of $768$.

\paragraph{Adapter Training and Optimization.} We train the argumentative adapters separately on Args.me and CMV for each of the models. Concretely, we train for $10$ epochs using the Adam optimizer~\citep{AdamW} (weight decay = $0.01$, $\beta_1 = 0.9$, $\beta_2 = 0.999$, $\epsilon = 1 \cdot 10^{-6}$, learning rate=$1\cdot 10^{-4}$) and early stopping based on the  perplexity on the validation set (patience: $2$ epochs). We set the effective batch size to $32$ except for training DialoGPT, for which we employ an effective training batch size of $8$ for reasons of computational capacity. The adapter reduction factor is $16$. %

\begin{figure}[t]
\begin{subfigure}{.5\textwidth}
  \centering
  \includegraphics[width=.9\linewidth]{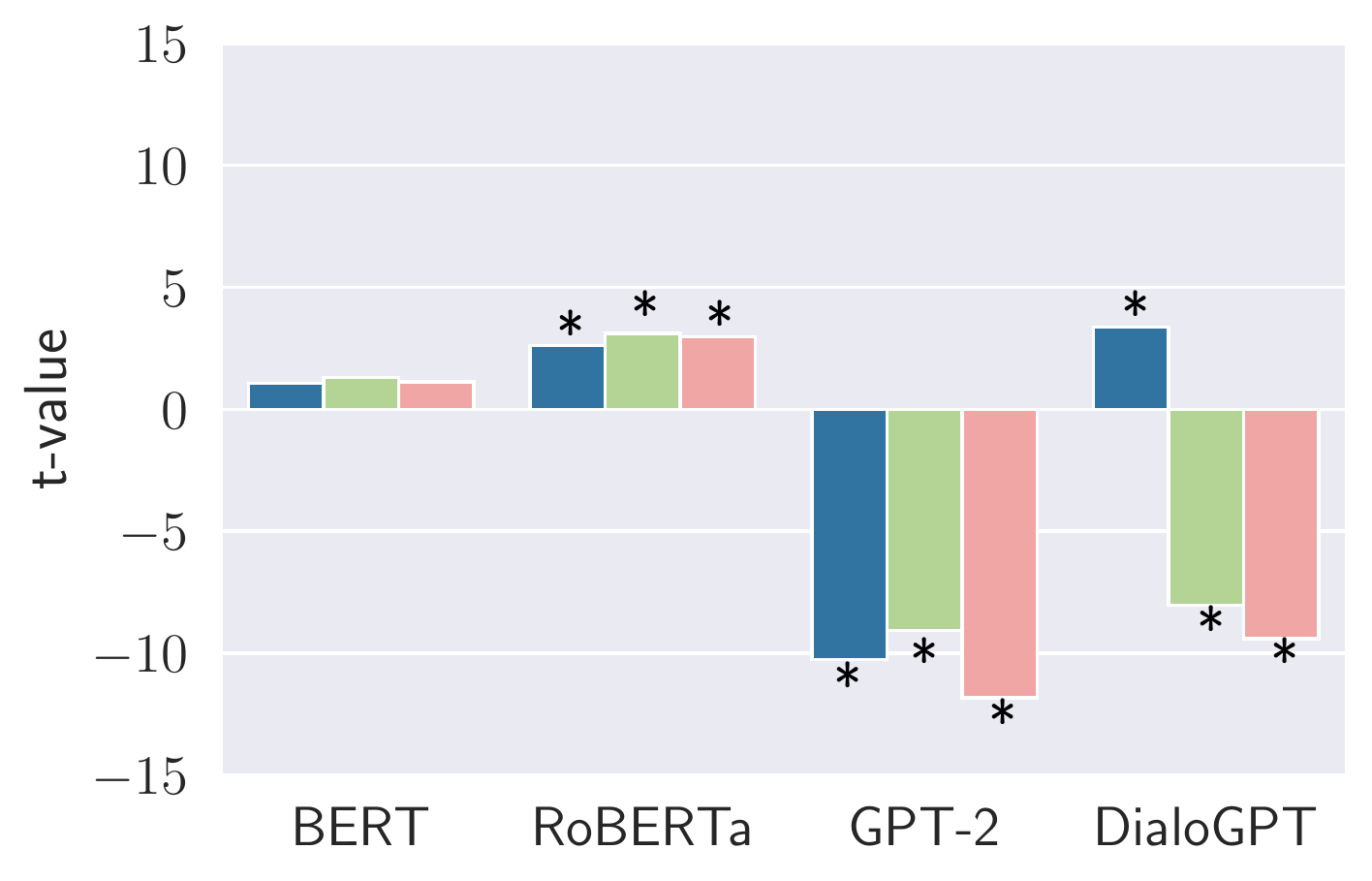}  
  \caption{LMB for Queerphobia}
  \label{fig:tt_queer}
\end{subfigure}
\begin{subfigure}{.5\textwidth}
  \centering
  \includegraphics[width=.9\linewidth]{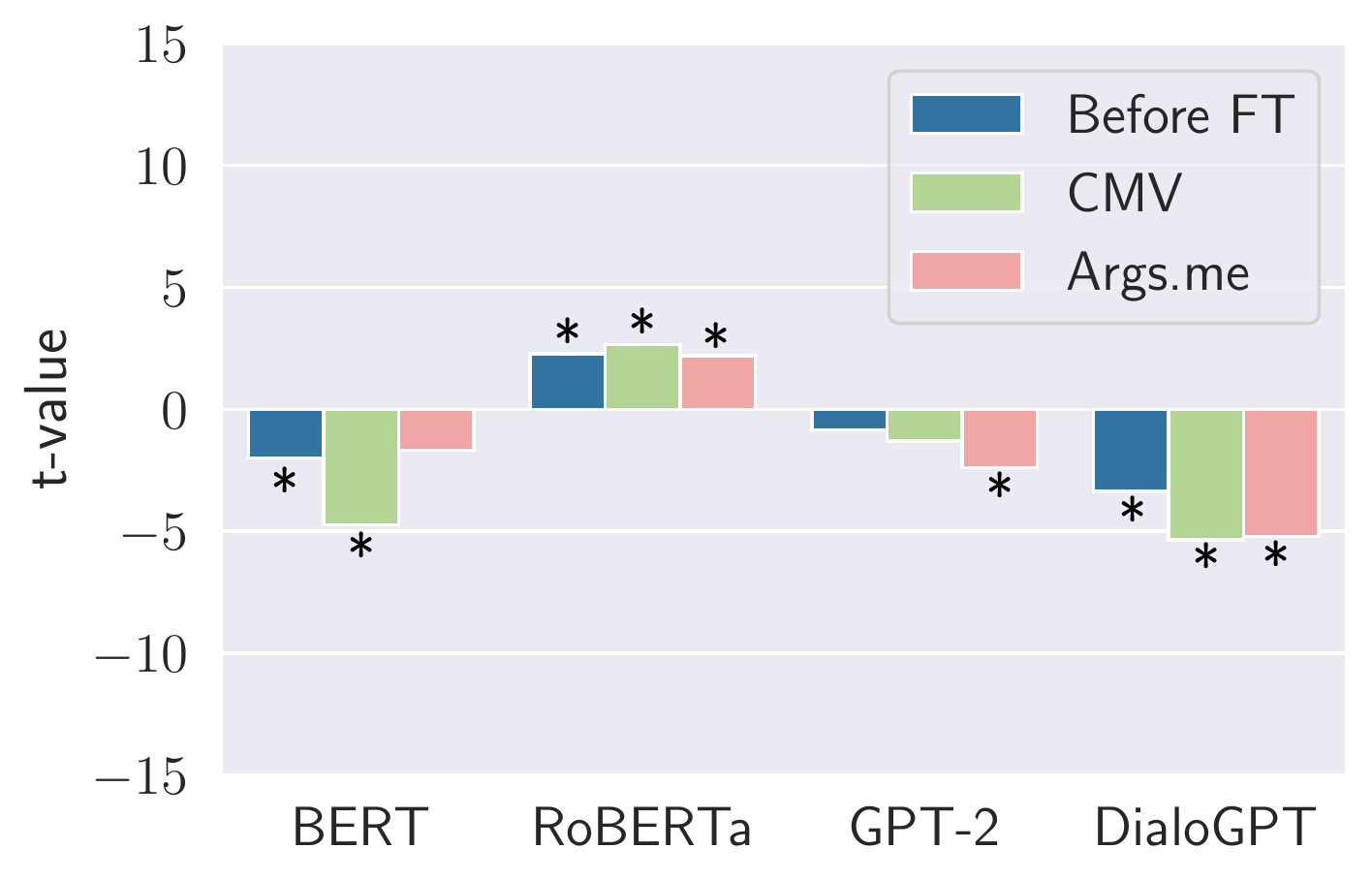} 
  \caption{LMB for Islamophobia}
  \label{fig:tt_rel}
  
\end{subfigure}

\caption{LMB scores before (\emph{Before FT}) and after argumentative fine-tuning on CMV and Args.me, respectively. Negative t-values indicate stereotypical biases. We highlight significant effect sizes with asterisks.}\label{fig:ALMResults}
\end{figure}

\paragraph{Results.} The LMB scores on \corpus before and after fine-tuning the four PLMs are shown in Figure~\ref{fig:ALMResults}. %
A negative t-value suggests a stereotypical bias; a positive t-value denotes an counter-stereotypical LMB, respectively.

\setlength{\tabcolsep}{2pt}
\begin{table}[t]
    \centering
    \small{
    \begin{tabular}{lr}
        \toprule
          \textbf{Sentence} & \textbf{P.} \\
        \midrule
         $S$: \,\textit{whats normal for gay people is immoral for us.} &  218 \\
        $S'$:\,\textit{whats normal for straight people is immoral for us.} &  363\\

        \bottomrule
    \end{tabular}}
    \caption{Example pair consisting of a biased ($S$) and inversely biased ($S'$) sentence exhibiting high difference in model perplexity (P.) for GPT-2 and Queerphobia.}
    \label{tab:perplSents-main}
\end{table}

Before fine-tuning, GPT-2 is the only model that exhibits a significant stereotypical bias along the \textit{Queerphobia} dimension. We show an example sentence pair exhibiting a high difference in model perplexity in Table~\ref{tab:perplSents-main} and provide more examples in the Appendix. %
For BERT, no significant difference was found between the perplexities on stereotypical and counter-stereotypical sentences along \textit{Queerphobia}, whereas RoBERTa and DialoGPT even show a significant counter-stereotypical bias. %
All PLMs except RoBERTa exhibit a stereotypical bias for the \textit{Islamophobia} bias, with a significant effect size for DialoGPT and BERT. The findings for DialoGPT are consistent with the results of \citet{barikeri2021redditbias} for conversational text.

When adapter-fine-tuning the PLMs on argumentative texts (CMV, Args.me), we notice that the perplexities on \corpus decreased, indicating that we successfully managed to inject argumentative knowledge into the models. However, we also observe that while for RoBERTa, no significant changes in t-values for either bias dimension occur, the sterotypical bias effects of DialoGPT and GPT-2 along the \textit{Islamophobia} bias dimension are reinforced by argumentative fine-tuning. 
Most interesting is the effect on DialoGPT along \textit{Queerphobia}. While the original model exhibited a significant counter-stereotypical bias, fine-tuning results in an opposite bias effect for both CMV and Args.me. Given that the stereotypical bias along the \textit{Islamophobia} dimension is also reinforced by fine-tuning DialoGPT, it underscores the tendency of the model to pick up and amplify stereotypical biases.
All in all, \emph{these findings highlight the importance of carefully measuring bias after injecting argumentative knowledge into the models.}

\subsection{Validating the Effectiveness of Adapter-based Debiasing}
\label{sec:rq2}

\paragraph{Debiasing Data.} 

\setlength{\tabcolsep}{3pt}
\begin{table}[t]
    \centering
    \small
    \begin{tabular}{llrrrr}
        \toprule
         & & \multicolumn{2}{c}{\textbf{Args.me}} & \multicolumn{2}{c}{\textbf{Wikipedia}} \\
         
        & \textbf{Strategy} & \textbf{\# Train} & \textbf{\#  Val.} & \textbf{\# Train} & \textbf{\# Val.}  \\
        \midrule
        \multirow{2}{*}{Q.}  & w/ N & 3,006,784 & 751,697 & 9,984,410& 2,496,103\\
         & w/o N & 80,598 & 20,150 & 43,616 & 10,904\\
        \midrule
        \multirow{2}{*}{I.}  & w/ N & 3,037,497 &  759,375 & 10,209,922 & 2,552,481 \\
         & w/o N & 142,024 & 35,506 & 494,640 & 123,660\\
        \bottomrule
    \end{tabular}
    \caption{Number of sentences in the training and validation portions of CDA-augmented Wikipedia and Args.me corpora. We report the sizes for Queerphobia (Q.) and Islamophobia (I.) and with (w/ N) and without neutral sentences (w/o N).}
    \label{tab:cdaDatasets}
\end{table}

We perform our two CDA strategies from \S\ref{sec:debiasingMethod} on two corpora: (i) the English Wikipedia ({\small\texttt{20200501.en}} dump) representing general-purpose encycopledic text. We randomly subsample the corpus, originally consisting of 6,078,422 text blocks, to 500,000 text blocks. (ii) We additionally experiment with the Args.me corpus, which also serves as the source for argumentative text.  %
On both corpora, we perform a train-validation slit of 80:20. 
The resulting train and test set sizes for both bias types \textit{Queerphobia} and \textit{Islamophobia} are listed in Table \ref{tab:cdaDatasets}.

\paragraph{Models.} We focus on two PLMs that exhibited bias along one of the dimensions in the previous experiments and which represent different types of PLMs: BERT as a representative of models trained via masked language modeling and GPT-2 as a model trained via causal language modeling.

\paragraph{Adapter Training and Optimization.} We train the adapters for $10$ epochs on the CDA-augmented data sets which include the neutral sentences, and for $1$ epoch on the data sets  that exclude the neutral sentences. The rest of the training procedure and all other hyperparameters are the same as for training the argumentaive adapters. %

\paragraph{Results.}

\begin{figure}[t]
\begin{subfigure}{.5\textwidth}
  \centering
  \includegraphics[width=.9\linewidth]{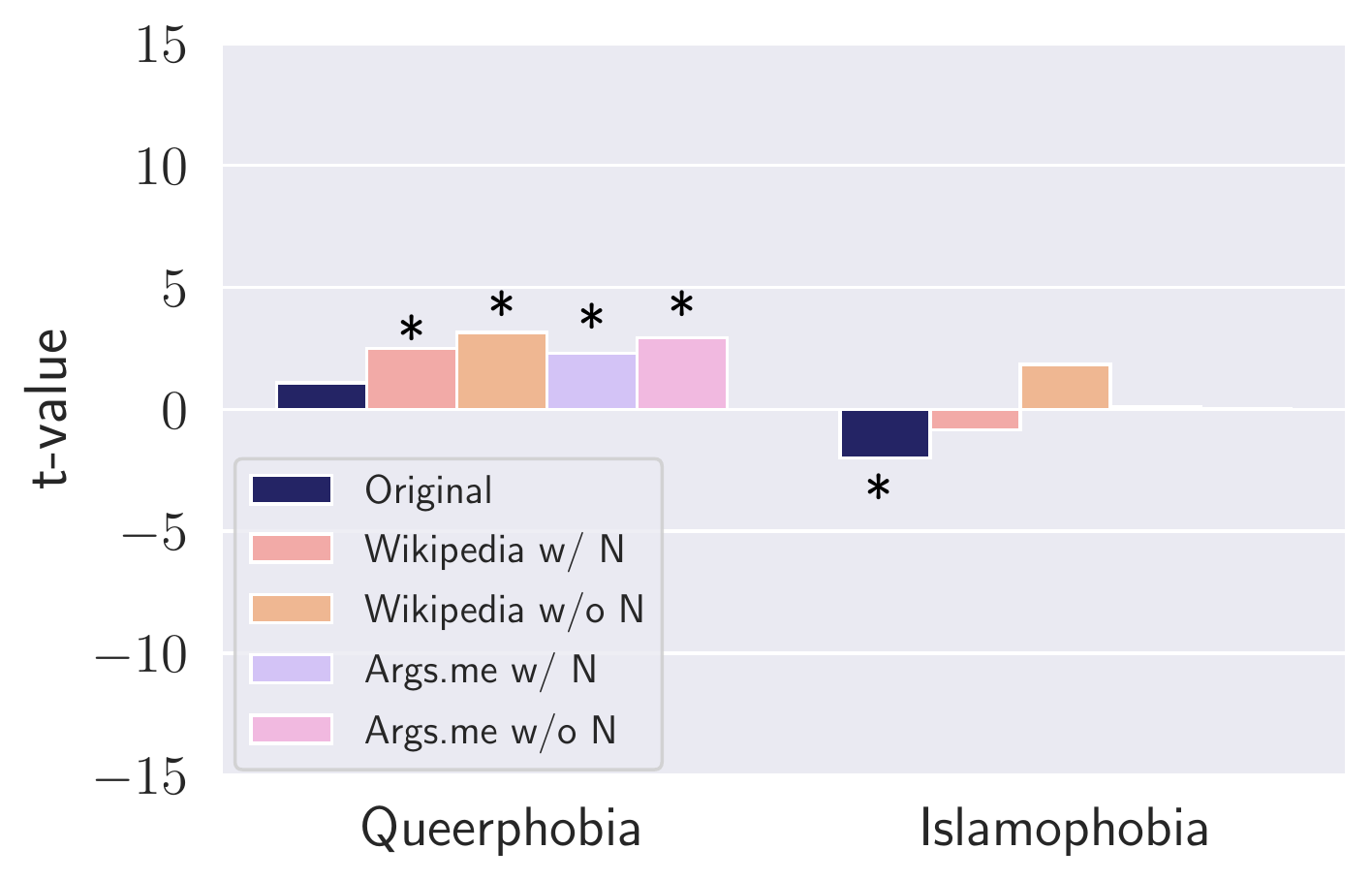}  
  \caption{BERT}
  \label{fig:sub-third}
\end{subfigure}
\begin{subfigure}{.5\textwidth}
  \centering
  \includegraphics[width=.9\linewidth]{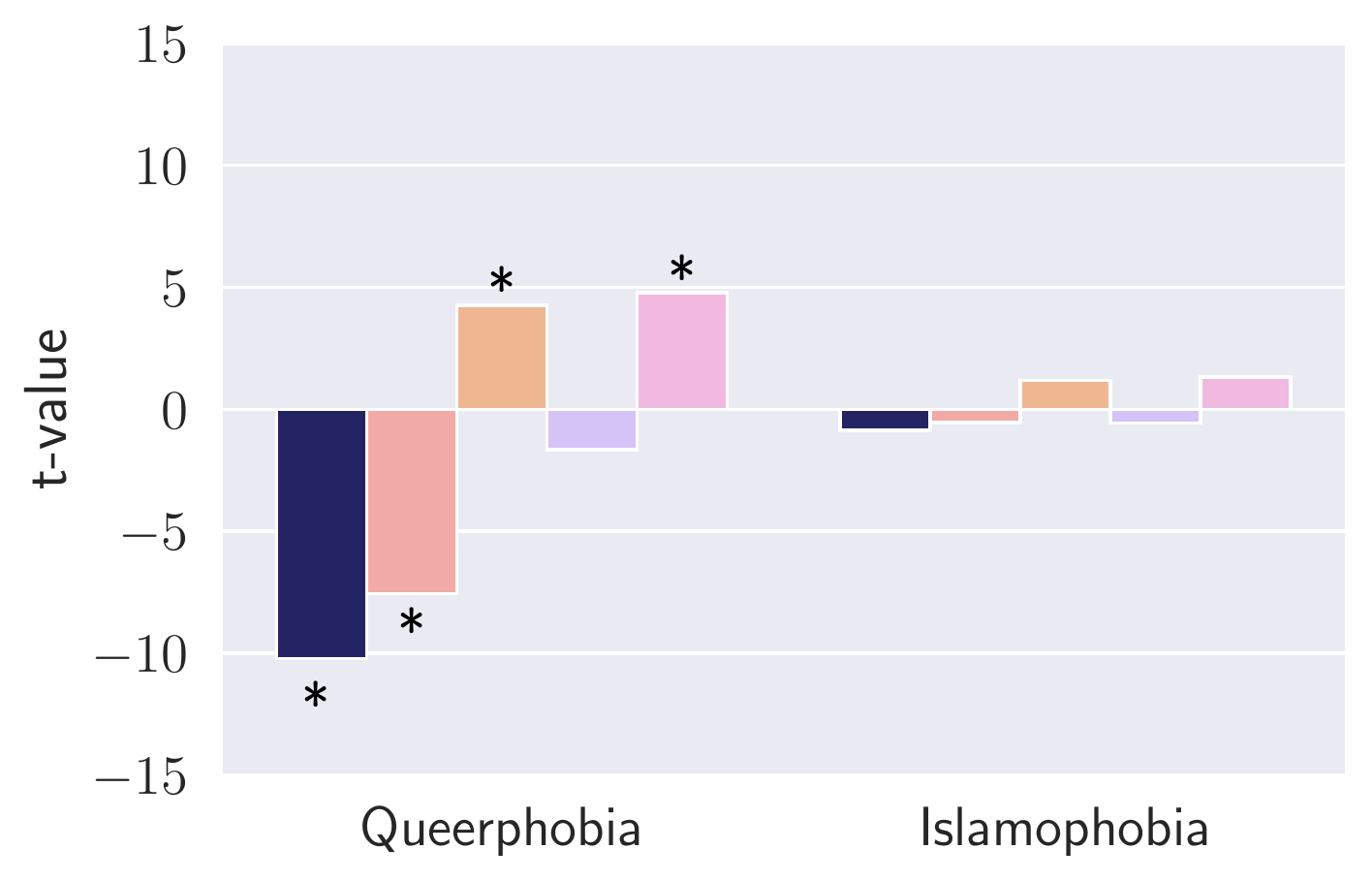}  
  \caption{GPT-2}
  \label{fig:sub-fourth}
\end{subfigure}
\caption{Debiasing results for BERT and GPT-2. We report LMB score (t-value) before and after injecting debiasing adapters trained on Wikipedia and Args.me with (w/ N) and without (w/o N) neutral sentences.}
\label{fig:DB}
\end{figure}

We report bias effect size using LMB in Figure \ref{fig:DB}. The results indicate that, while the original PLMs exhibited significant bias along a dimension, \emph{using debiasing adapters we are able to successfully reduce the measurable bias from a significant to a non-significant amount}, the only exception with the adapters for GPT-2 trained on the CDA-augmented Wikipedia. When we exclude neutral sentences the scores switch into the counter-stereotypical direction: we hypothesize that this indicates the need for a better balancing and sampling of the training data. We see a similar effect for cases in which the original PLM did not exhibit a significant bias -- the LMB is likely to switch to the opposite, counter-stereotypical direction.

\subsection{Combining Argumentative Knowledge and Fairness}
\label{sec:rq3}
Taking advantage of the modular nature of adapters, we combine argumentation and debiasing adapters (\S\ref{sec:rq1}-\ref{sec:rq2}) to obtain a fair and argumentative language model using \emph{AdapterStacking} (\S\ref{sec:combining}). We focus on the bias dimensions for which the original models exhibited a stereotypical effect size.

\paragraph{Results.} 
\begin{figure}[t]
\begin{subfigure}{.5\textwidth}
  \centering
  \includegraphics[width=.9\linewidth]{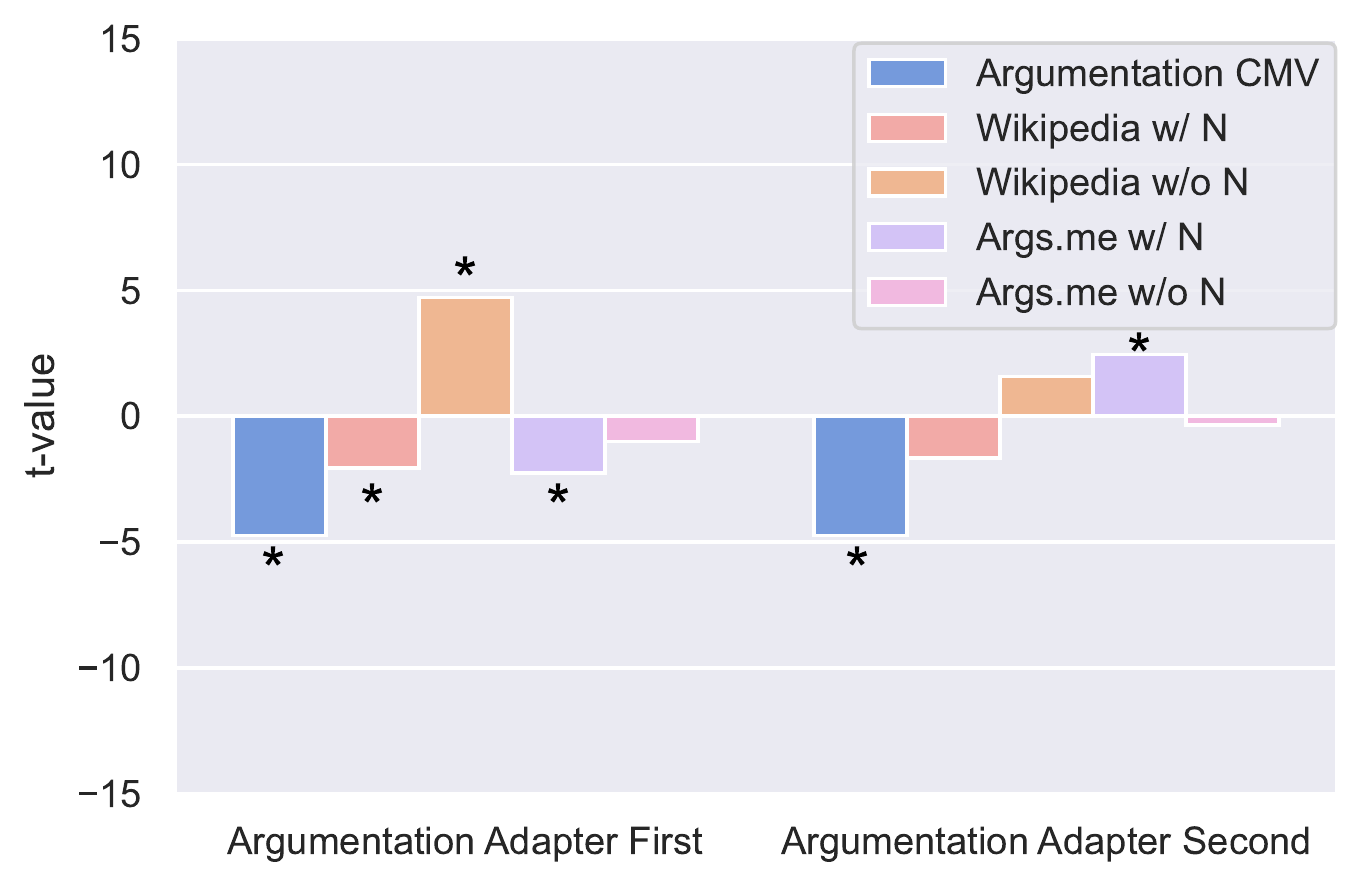}  
  \caption{Islamophobia LMB for BERT}
  \label{fig:stacking_results_bert}
\end{subfigure}
\begin{subfigure}{.5\textwidth}
  \centering
  \includegraphics[width=.9\linewidth]{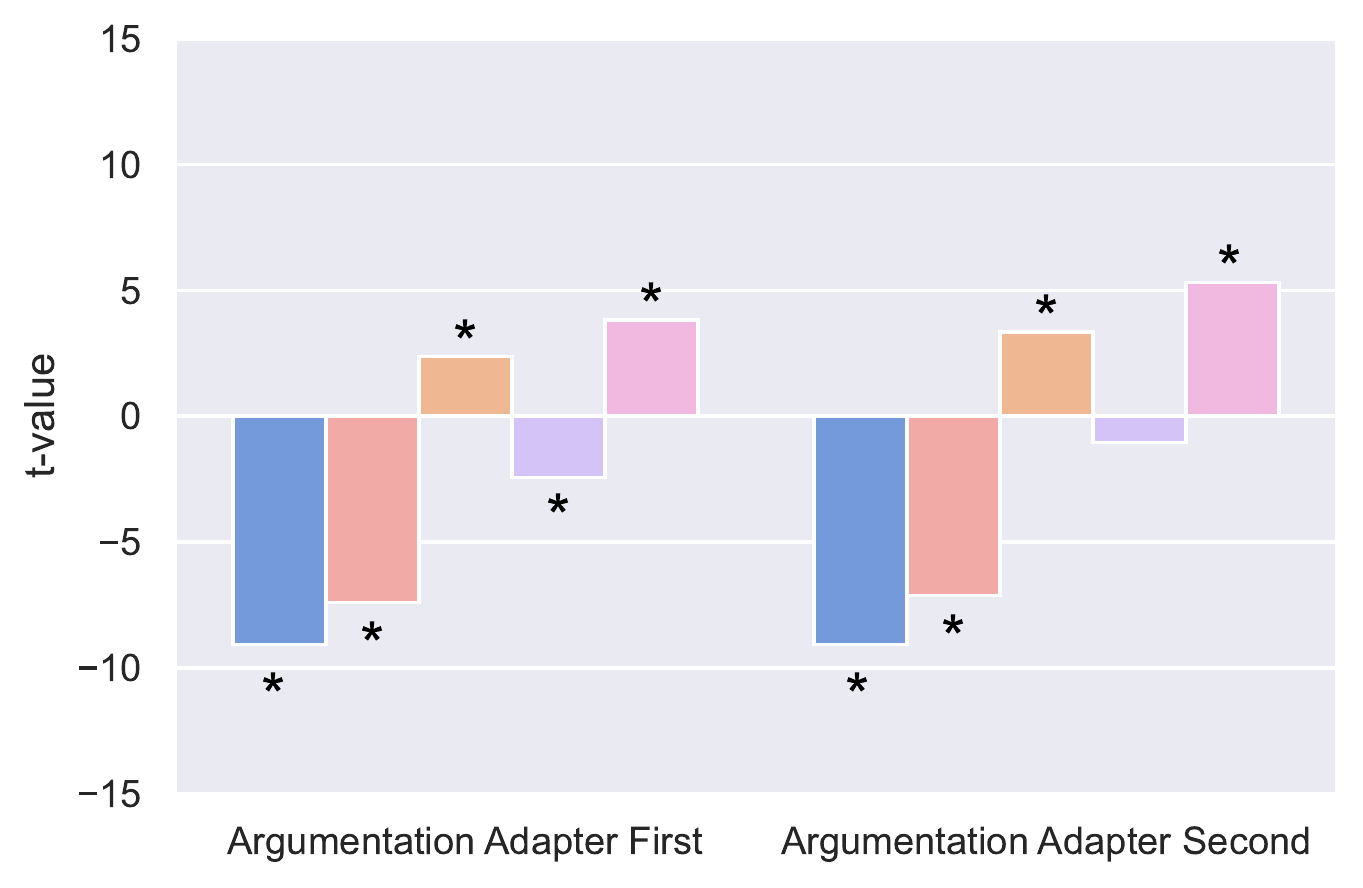} 
  \caption{Queerphobia LMB for GPT-2}
  \label{fig:stacking_results_gpt}
\end{subfigure}
\caption{LMB for different stacking orders of the argumentation adapter (left: \emph{argumentation adapter} first; right: \emph{debiasing adapter} first).}
\label{fig:Stacking_results}
\end{figure}

Figure \ref{fig:Stacking_results} shows the LMB scores of BERT on Islamophobia and GPT-2 along Queerphobia for different stacking orders of the argumentation adapter trained on CMV and the respective debiasing adapters trained on Wikipedia or Args.me (results for the other dimensions and other argumentation adapters are found in the Appendix).  For BERT, stacking the debiasing adapters for Islamophobia second and the argumentation adapter trained on CMV first (left) reduces the bias to an non-significant amount only in a single case, while stacking the debiasing adapter first (right) removes the bias in three out of four setups. Also for GPT-2, stacking the debiasing adapter first leads to better debiasing results.
We hypothesize that the reason for this effect is that both types of adapters are optimized for receiving the input directly from the transformer layers. Thus, the debiasing adapter is more effective when stacked first. In sum, while our results indicate that \emph{stacking order matters and debiasing effects are bigger when debiasing adapters are stacked first}, we think that this finding warrants future research on the issue.

\subsection{Downstream Evaluation on Argument Quality Prediction}
\label{sec:rq4}

\setlength{\tabcolsep}{3.5pt}
\begin{table}[t]
    \centering
    \small
    \begin{tabular}{llccc}
        \toprule
        \textbf{Dataset} & \textbf{Domain} & \textbf{\# Train} &  \textbf{\# Validation} & \textbf{\# Test}\\ 
        \midrule
        IBM-Rank-30k & -- & 20,974 & 3,208 & 6,315\\
        \multirow{3}{*}{GAQCorpus} & CQA & 1,109 & 476 & 500\\
        & Debates & 1,093 & 469 & 538\\
        & Reviews & 700 & 400 & 100\\
        \bottomrule
    \end{tabular}
    \caption{Number of arguments in training, validation, and test portions of IBM-Rank-30k and GAQCorpus.}
    \label{tab:AQdata}
    \vspace{-1.5em}
\end{table}

\paragraph{Data and Measures.} For testing the influence of our argumentation and debiasing adapters on argument quality prediction, we employ two recently presented data sets: (1) the IBM-Rank-30k~\citep{LargeScaleDSArgument}, an extension of  \citep{toledo-etal-2019-automatic}, which consists of short-length arguments (maximum length of 210 characters) annotated by crowd workers. We use the MACE-P aggregations provided by the authors for model training. (2) Additionally, we use the GAQCorpus~\citep{GAQCorpus, lauscher-etal-2020-rhetoric} which covers real-world arguments from three domains, namely community questions and answers (CQA), online debate forums (Debates), and restaurant reviews (Reviews). An overview of the data sets is given in Table~\ref{tab:AQdata}. On both data sets, we report Pearson's correlation coefficient ($r$). %
Following \citet{ReportingScore}, we report the average of our experiments conducted $50$ times with different random seeds (using the best hyperparameter configuration according to the development set results) and additionally conduct an independent t-test. %

\paragraph{Models.} For all AQ models, we rely on a simple linear regression head into which we input the pooled sequence representation. The fine-tuning strategy for the AQ regression is aligned with our previous approaches. Instead of full fine-tuning of the encoder, we add an additional task-specific adapter on top of the already existing adapters and adjust only the task-specific adapter parameters during training. As before, we employ the BERT and GPT-2 base models (\texttt{Base}) as well as the adapter-augmented variants. Concretely, we employ the argumentation adapters trained on Args.me and CMV (\texttt{Argsme}, \texttt{CMV}), and the debiasing adapters trained on the CDA-augmented Args.me (\texttt{DB-Islamo} for BERT, \texttt{DB-Queer} for GPT-2). Again, we also study combinations to optimally combine argumentation, debiasing, and task-specific knowledge using either a stacking  (\texttt{Stacked}) or fusion architecture (\texttt{Fusion}).  On IBM-Rank-30k, we follow \citet{LargeScaleDSArgument} and concatenate topic and argument with an additional separator (BERT) or end-of-sequence token (GPT-2).
As baselines, we additionally compare with the best results reported by the original works. %

\paragraph{Adapter Training and Optimization.}  Following \citet{LargeScaleDSArgument} and \citet{lauscher-etal-2020-rhetoric}, we optimize our models using Mean Squared Error. We train all task adapters using Adam~\citep{AdamW} with a batch size of $32$ (weight decay = $0$, $\beta_1 = 0.9$ and $\beta_2 = 0.999$). We pad the input sequences to a maximum length of 128. %
We choose the best hyper-parameters by grid searching for learning rate $\lambda \in \{1 \cdot 10^{-4},2\cdot 10^{-4},3\cdot 10^{-4}\}$ and number of training epochs $\in \{1,2,3,4,5\}$ based on the performance on the individual dataset's respective validation portion.  %

\paragraph{Results.} 
\setlength{\tabcolsep}{2.5pt}
\begin{table}[t]
    \centering
    \small
    \begin{tabular}{lp{2.5cm}llll}
        \toprule
        && \textbf{IBM} &  \multicolumn{3}{c}{\textbf{GAQ}}\\ %
        &\textbf{Model} &   &  \textbf{CQA} & \textbf{Debates} &  \textbf{Reviews} \\%& \textbf{CQA} & \textbf{Deb.} &  \textbf{Rev.}\\
        \midrule
        \multicolumn{2}{l}{\citet{LargeScaleDSArgument} } & 0.53  & & & \\%& 0.676 & 0.545 & 0.596\\
        \multicolumn{3}{l}{\citet{lauscher-etal-2020-rhetoric}}  & 0.652 & 0.511 & 0.605 \\%& 0.676 & 0.545 & 0.596\\
        \midrule
        \parbox[t]{2mm}{\multirow{6}{*}{\rotatebox[origin=c]{90}{BERT}}} & \texttt{Base} & 0.524   & 0.663 & 0.465 & 0.560\\  %
        &\texttt{Argsme} & 0.531*  & 0.600* & 0.439* & 0.511* \\%& 0.636* & 0.491* & 0.637\\
        &\texttt{CMV} & 0.525 &  0.608* & 0.453 & 0.521* \\%& 0.671* & 0.502* & 0.630\\
        &\texttt{DB-Islamo} & \textbf{0.531*}  & 0.653* & 0.479* & 0.560 \\%& 0.676* & 0.508* & 0.630\\
        &\texttt{Stacked} & 0.528*  & 0.663 & \textbf{0.485*} & 0.528* \\%& 0.672* & 0.519 & 0.639*\\
        &\texttt{Fusion} &  0.521*  & \textbf{0.672*} & \textbf{0.487*} & \textbf{0.569*} \\%& 0.680* & 0.518 & \textbf{0.641*}\\
    \midrule
        \parbox[t]{2mm}{\multirow{6}{*}{\rotatebox[origin=c]{90}{GTP-2}}} & \texttt{Base} & 0.513  & 0.658 & 0.474 & 0.519 \\%& \textbf{0.685} & \textbf{0.518} & 0.563\\
        &\texttt{Argsme} & 0.512  & 0.612* & 0.407* & 0.496 \\%& 0.662* & 0.493* & 0.574\\
        &\texttt{CMV} & \textbf{0.516*}  & 0.626* & 0.419* & 0.504 \\%& 0.672* & 0.510 & 0.592* \\
        &\texttt{DB-Queer} & 0.512 &   0.62* & 0.476 & 0.507\\%& 0.671* & 0.514 & 0.561\\
        &\texttt{Stacked} & 0.513  & 0.609* & 0.428* & 0.515\\%& 0.653* & 0.488* & \textbf{0.581*}\\
        &\texttt{Fusion} & 0.507* & \textbf{0.683*} & \textbf{0.488*} & \textbf{0.528} \\%& \textbf{0.685} & 0.517 & 0.576* \\
        \bottomrule
    \end{tabular}
    \caption{Argument Quality prediction results (mean Pearson's correlation across 50 runs) on IBM-ArgQ-Rank-30kArgs and GAQCorpus. (*) indicates statistically significant differences.}
    \label{tab:AQresults}
    \vspace{-1em}
\end{table}

The results are shown in Table~\ref{tab:AQresults}. Generally, though the trends are the same, the scores diverge from the results reported in the original works, which can be attributed to our use of task adapters. Interestingly, while injecting argumentation adapters leads to performance improvements on IBM-ArgQ-Rank-30kArgs in 3 out of 4 cases, it seems to hurt the performance on GAQCorpus. On the other hand, the debiasing adapters do not seem to lead to losses: in contrast, in some cases (IBM and GAQ--Debates for BERT, GAQ--Debates for GPT-2), we even note performance improvements. For GAQCorpus, the best results are obtained with an argumentative and fair language model -- when fusing debiasing and argumentation adapters. We conclude that \emph{fair and argumentative language modeling can have a positive impact on argument quality prediction as downstream task}.

%% file: 06-rw.tex
\paragraph{Bias in NLP.} For thorough reviews on bias mitigation and evaluation we refer to \citet{blodgett-etal-2020-language}, and \citet{shah-etal-2020-predictive}. \citet{Bolukbasi} were the first to draw attention to the issue of unfair stereotypical bias in NLP, showing that static word embeddings allow for building biased analogies. Later, \citet{Caliskan_2017} proposed the well-known Word Embedding Association Test (WEAT), which was extended to more languages by \citep{lauscher-glavas-2019-consistently,lauscher-etal-2020-araweat}. More works focused on bias evaluation and mitigation in static word embeddings~\citep{gonen-goldberg-2019-lipstick,dev2019attenuating, manzini-etal-2019-black, DEBIE}, and later, the focused shifted towards detecting and attenuating biases in their successors contextualized word embeddings~\citep{dev2019attenuating, dev2020measuring, ABSIntersectional}.
Here, the authors focused on both, bias in general-purpose pretrained language models~\citep{may-etal-2019-measuring, MeasuringBIASBERT, DebiasingELMO, webster2020measuring}, and bias in particular downstream scenarios~\citep{dev2020measuring}. For instance, \citet{zhao_etal_2018_cda} proposed Counterfactual Data Augmentation (CDA) for the purpose of debiasing coreference resolution systems. Like many other works~\citep{zmigrod-etal-2019-counterfactual,Lu2020,webster2020measuring,lauscher2021sustainable} we explore the method for our purposes. Similarly, \citet{vanmassenhove-etal-2018-getting} focused on machine translation and \citet{thewomanworked} on general natural language generation, while \citet{barikeri2021redditbias} specifically target conversational models. In this work, we follow their process for creating \corpus.

\paragraph{Bias in Argumentation.} It is extremely surprising that given the plethora of works focused on mining, assessing, and generating arguments as well as reasoning over arguments \citep{lauscher2021scientia}, to date, \citet{spliethoever} were the only ones to investigate and quantify social bias in argumentation. They performed a simple co-occurrence analysis for three different argumentation corpora and trained a custom GloVe model~\citep{glove} based on argumentative text, which they analyzed with WEAT. Our work builds on top of theirs and is the first to examine bias in relation to an argumentative downstream task and also the first to conduct debiasing for computational argumentation models. %

%% file: 07-conclusion.tex
In this work, we presented an investigation of bias in PLMs and argumentative text. To this end, we created \corpus, the first annotated corpus tailored for measuring bias in computational argumentation models. Using \corpus, we showed that argumentative fine-tuning of language models may lead to an amplification of biases in the models. We then demonstrated how to obtain a fair and argumentative language model by combining argumentation with debiasing knowledge encapsulated in lightweight adapters to ensure higher sustainability and flexibility, and analyzed the effect of stacking orders.  An additional downstream evaluation on argument quality prediction indicated that debiasing can even lead in some cases to improved results. We hope that with this work, especially the novel \corpus resource, we will foster further research on fair computational argumentation.

%% file: 08-limitations.tex
We like to point the reader to the following limitations and ethical considerations: first, following the large body of debiasing research in NLP, we based our evaluation, mitigation, and annotation approach on a fixed set of manually created terms. We are aware that this set is never finite and may be continually revised in subsequent studies. For a recent discussion we refer to \citet{antoniak-mimno-2021-bad}. This is especially the case for the dimension of \textit{Queerphobia}, where there is increasing openness and understanding toward more diverse forms of sexual orientation and (gender) identity. For instance, our vocabulary does not include the variety of gender-neutral (neo)pronouns~\citep[][]{dev-etal-2021-harms, lauscher2022welcome}.  %
Further, studies have shown that the perception of prejudice is not only highly subjective, but also largely culture-dependent \citep{webster2020measuring}.  Consequently, in order to conduct a thoroughly unbiased annotation study, annotators should be carefully selected and as diverse as possible in terms of cultural heritage, age, ethnicity, and religious affiliation, as well as their gender identity and sexual orientation. While our three annotators were of diverse cultural background such diversity of human resources was not available for this work.

%% file: xx-appendix.tex
\section{Bias Specifications}\label{app:targetTerms}
We provide the full sets of target and attribute terms for Queerphobia and Islamophobia.

\subsection{Queerphobia}

\paragraph{List of Target Terms $\mathbf{T_1}$:} gay, gays, lesbian, lesbians, bisexual, bisexuals, homosexual, homosexuals, transgender, transgenders, sapphic, sapphics, pansexual, pansexuals, queer, queers, genderqueer, genderqueers, transsexual, transsexuals, crosssexual, crosssexuals, epicene, epicenes, lgbtq, hermaphroditic, hermaphrodite,\\ hermaphrodites, androgynous, androgyne, androgynes, intersex

\paragraph{List of Target Terms $\mathbf{T_2}$:} straight, straights, monosexual, monosexuals, heterosexual, heterosexuals, cisgender,
                cisgenders, sexually normal, hetero, heteros, unisexual, unisexuals

\paragraph{List of Attribute Terms $\mathbf{A_1}$:} mentally ill, polygamous, flamboyant, pedophile, sexual predator, mental disorder,
                    irreligious, villain, serial killer, depress, adultery, drug addict, sadist, arts,
                    artistic, theatre, dramatic, show queen, fashion, predator, party drugs,
                    illicit drugs, alcohol, sexual abuse, child rapist, indecisi, promiscuous,
                    sexually manipulative, prostitute, drag queen, drag king, criminal, hypersexual,
                    asexual, pervert, perverts, poofter, faggot, fag, weak, timid, loser, loner,
                    lonely, sin, disease, confuse, strange, weird, coward, sexually abnormal,
                    immoral,
                    insecure, repulsive, frustrat, sinful, sensitive, weak minded, lack self control,
                    oversex, submissive, emotional

\paragraph{List of Attribute Terms $\mathbf{A_2}$:} mentally, strong, monogamous, modest, teleiophile, protector, mental stability,
                    hero, police officer, cheerful, healthy, compassionate, scientific, science, subtle,
                    unfashion, defender, determined, moral, chaste, cisgender, lawful, decent, saint,
                    faggot, strong, daring, winner, cheerful, virtue, health, assured, normal,
                    normal, brave, moral, confident, delightful, satisfied, innocent, tough,
                    dominating, unemotional

\begin{table*}[t]
    \centering
    \begin{tabular}{c c c}
    \toprule
    \textbf{Sexual Identity} &  \textbf{Gender Identity} &  \textbf{Biological Sex}\\
    \midrule
    (gay, heterosexual) & (transgender, cisgender) & (epicene, unisexual) \\
    (gays, heterosexuals) & (transgenders, cisgenders) & (epicenes, unisexuals) \\
    (lesbian, heterosexual) & (transsexual, cisgender) & (hermaphroditic, unisexual) \\
    (lesbians, heterosexuals) & (transsexuals, cisgenders) & (hermaphrodite, unisexual) \\
    (sapphic, heterosexual) & (crosssexual, cisgender) & (hermaphrodites, unisexuals) \\
    (sapphics, heterosexuals) & (crosssexuals, cisgenders) & (androgynous, unisexual) \\
    (homosexual, heterosexual) & (genderqueer, cisgender) & (androgyne, unisexual) \\
    (homosexuals,heterosexuals) & (genderqueers, cisgenders) & (androgynes, unisexuals) \\
    (queer, heterosexual) & (queer, cisgender) & (intersex, unisexual) \\
    (queers, heterosexuals) & (queer, cisgenders) &  \\
    (lgbtq, heterosexual) & (androgyne, cisgender) &  \\
    (gay, straight) & (androgynes, cisgenders) & \\
    (gays, straights) &  &  \\
    (lesbian, straight) & & \\
    (lesbians, straights) & & \\
    (sapphic, straight) & & \\
    (sapphics, straights) & & \\
    (bisexual, monosexual) & & \\
    (bisexuals, monosexuals) & & \\
    (pansexual, monosexual) & & \\
    (pansexuals, monosexuals) & & \\
    \bottomrule
    \end{tabular}

    \caption[Target Term Pairs for Queerphobia]{Target Term Pairs for Queerphobia}
        \label{tab:pairs_queer}
\end{table*}

The target term pairs used are depicted in Table~\ref{tab:pairs_queer}.

\subsection{Islamophobia}

\paragraph{List of Target Terms $\mathbf{T_1}$:} muslim, muslims, islam, islamic, islamism, arab, arabs, quran, koran

\paragraph{List of Target Terms $\mathbf{T_2}$:} christian, christians, christianity, church, bible, priest

\paragraph{List of Attribute Terms $\mathbf{A_1}$:} terrorist, terrorism, threat, threaten, dangerous, deadly, attack, bomb, murder,
                    hijack, criminal, enemy, war, enemies, extremist, radical, hate, rich, lazy,
                    flashy, lazy sheik, oil sheik, rapist, slaves, shooter, belly dancers, hijab,
                    viel, vielded, oppressed, patriarchy, misogyny, patriarchal, deport, detain,
                    thieves, thug, charlatan, brute, power hungry, beard, wealthy oilmen,
                    harem maiden, suicide bomb, headscarves, violence, violent, hostile, sinister

\paragraph{List of Attribute Terms $\mathbf{A_2}$:} police, nonviolence, safety, defend, helpful, safe, defend, build, restore,
                    relinquish, lawful, friend, peace, friends, moderate, conservative, love, poor,
                    energetic, simple, defender, freemen, hero, hat, unviel, unvielded, rejoiced,
                    matriarchy, philogyny, matriarchal, admit, liberate, police, hero, honest, mild,
                    gratified, clean shave, negotiator, compassion, gentle, kind, happy

\begin{table}[t]
    \centering
    \begin{tabular}{c}
    \toprule
   \textbf{Islamophobia} \\
    \midrule
(muslim, christian)\\
(muslims, christians)\\
(islam, christianity)\\
(islamic, christian)\\
(islamism, christianity)\\
(arab, american)\\
(arabs, americans)\\
(koran, bible)\\
(quran, bible)\\
(imam, priest)\\

    \bottomrule
    \end{tabular}
    
    \caption[Target Term Pairs for Islamophobia]{Target Term Pairs for Islamophobia}
    \label{tab:pairs_rel}
\end{table}
The target term pairs used are depicted in Table~\ref{tab:pairs_rel}.

\section{Model Perplexity Examples}
Some examples of sentences where the models exhibit a large difference in perplexity are shown in Table \ref{tab:perplSents}. It is striking that although the sentence pairs are both grammatically correct, the perplexities calculated by the models differ greatly.

\setlength{\tabcolsep}{7.8pt}
\begin{table*}[t]
    \centering
    \begin{tabular}{lllr}
        \toprule
        \textbf{Model} & \textbf{Type} & \textbf{Sentences} & \textbf{Perplexity} \\
        \midrule
        \multirow{2}{*}{GPT-2} & \multirow{2}{*}{Queerphobia} & $S$ : \textit{whats normal for gay people is immoral for us.} &  218\\
        & & $S'$: \textit{whats normal for straight people is immoral for us.} &  363\\
        \hline
        \multirow{2}{*}{DialoGPT} & \multirow{2}{*}{Islamophobia} & $S$ : \textit{muslims follow the koran.. terrorism} &  296\\
        & &$S'$: \textit{christians follow the bible.. terrorism} & 977 \\
        \hline
         \multirow{2}{*}{BERT} & \multirow{2}{*}{Islamophobia} &$S$ : \textit{islam is murder.} &  1262\\
        & &$S'$: \textit{christianity is murder.} & 2054\\
\hline
         \multirow{2}{*}{RoBERTa} & \multirow{2}{*}{Queerphobia} &$S$ : \textit{transgender people are mentally ill.} &  2579\\
        & &$S'$: \textit{cisgender people are mentally ill.} & 926 \\

        \bottomrule
    \end{tabular}
    \caption{Examples of biased and inversely biased sentences exhibiting high differences in model perplexity.}
    \label{tab:perplSents}
\end{table*}

\section{Links to Datasets and Models}
\begin{table*}
\centering
{%
\begin{tabularx}{\textwidth}{l l X}
\toprule
{\bf Codebase} & {\bf Model}  & {\bf URL} \\ \midrule
Adapters & -- & \url{https://github.com/Adapter-Hub/adapter-transformers} \\ 
Transformers & -- &  \url{https://github.com/huggingface/transformers} \\ 
& BERT & {\url{https://huggingface.co/bert-base-uncased}} \\
& GPT-2 &  {\url{https://huggingface.co/gpt2}} \\
& DialoGPT &  {\url{https://huggingface.co/microsoft/DialoGPT-medium}} \\
& RoBERTa & {\url{https://huggingface.co/roberta-base}} \\
\bottomrule
\end{tabularx}
}
\caption{Links to codebases and pretrained models used in this work.}
\label{tbl:deps_models}
\end{table*}

\setlength{\tabcolsep}{6pt}
\begin{table*}%
\centering
\begin{tabularx}{\textwidth}{l l X}
\toprule
{\bf Purpose} & {\bf Dataset} & {\bf URL} \\ \midrule
Argument Quality  & GAQCorpus & \url{https://github.com/grammarly/gaqcorpus} \\
& IBM-Rank-30k & \url{https://research.ibm.com/haifa/dept/vst/debating_data.shtml\#Argument\%20Quality} \\
Argumentative LM & Args.me & \url{https://webis.de/data/args-me-corpus.html}\\
 & Webis-ChangeMyView-20 & \url{https://zenodo.org/record/3778298\#.YY5aLS9Q2J8} \\
CDA Debiasing & Wikipedia & \url{https://dumps.wikimedia.org/} \\
 & Args.me & \url{https://webis.de/data/args-me-corpus.html} \\
\bottomrule
\end{tabularx}
\caption{Links to the datasets used in our work. }
\label{tbl:datasets}
\end{table*}
We provide links to data sets, code bases, and all pretrained models used in this work in Tables~\ref{tbl:deps_models} and~\ref{tbl:datasets}.

\section{Further Experimental Results}
\begin{figure}[t]
\begin{subfigure}{.5\textwidth}
  \centering
  \includegraphics[width=\linewidth]{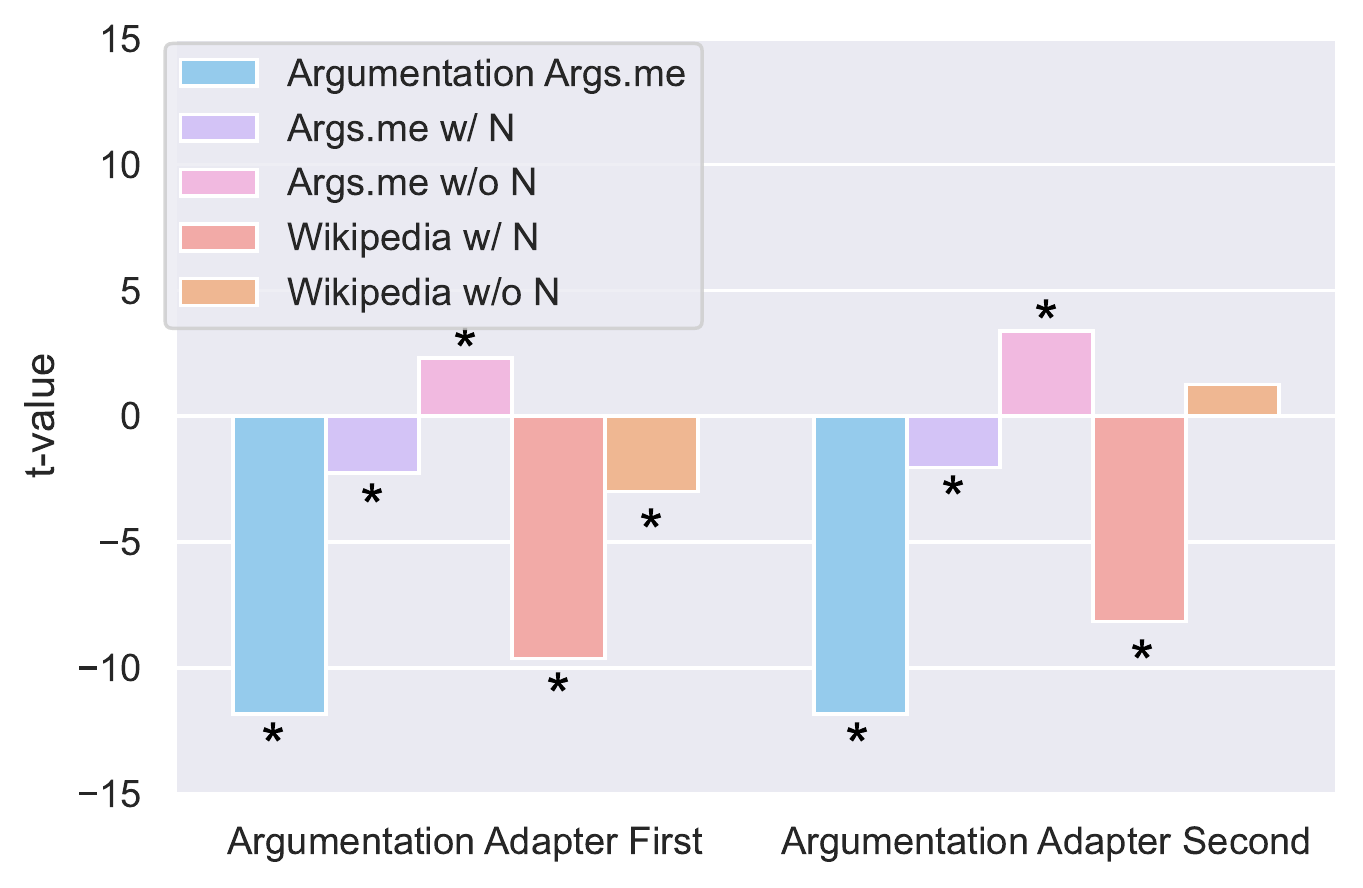}  
  \caption{Queerphobia}
  \label{fig:stack_gpt2_argsme_qp}
\end{subfigure}
\begin{subfigure}{.5\textwidth}
  \centering
  \includegraphics[width=\linewidth]{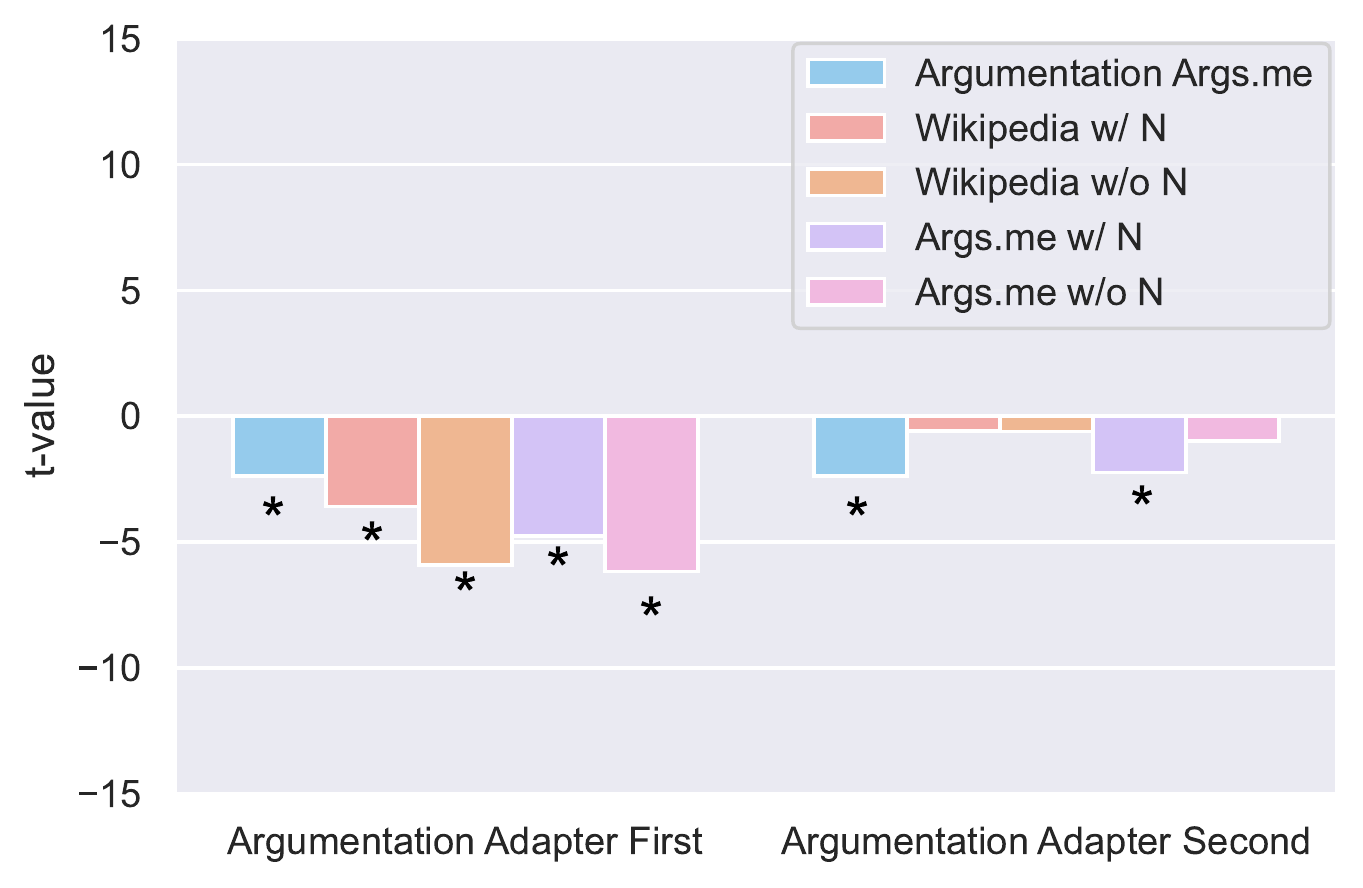}  
  \caption{Islamophobia}
  \label{fig:stack_gpt2_ip}
\end{subfigure}
\caption{LMB results for GPT-2 with the argumentative adapter trained on Args.me and respective stacking variants.}
\label{fig:StackGPTAdditional}
\end{figure}
We show the additional experimental results relating to RQ3 in Figure~\ref{fig:StackGPTAdditional}.

\section{Annotation Guidelines}
We provide the annotation guidelines we employed for creating \corpus.
\clearpage



\section*{Supplementary material}

# Debate.org Queerphobia Annotation Guidelines
# Version 2.0.0

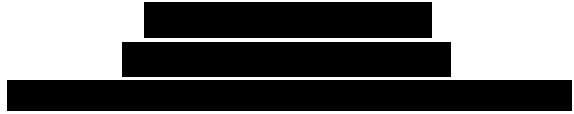

## 1 Introduction

Debate.org is an online debate portal that provides a platform for open discussion, where all members of the community can express their arguments on a wide range of controversial topics. This document describes the annotation guidelines for declaring these user arguments as either expressing stereotyped human bias or not.

According to the Oxford Dictionary of Psychology, a stereotype is defined as:

**"A relatively fixed and oversimplified generalization about a group or class of people."**
(Colman, 2015)

In this annotation task, we explicitly focus on the classification of negative or unfavorable biases.

For this purpose, a CSV file is provided that contains the user arguments, potentially biased sentences from these arguments, and the contained term pairs consisting of a target term referring to a demographic minority group (e.g., "homosexuals") and an attribute term referring to a presumably biased expression (e.g., "sin"). Based on this file, it is up to you to decide whether these extracted sentences or even the entire argument are indeed *biased* or *unbiased*.

If there are open questions that you feel are not covered in this guide, please do not hesitate to get in touch so that these questions can be discussed and the guide improved.

## 2 CSV Presentation

The CSV file you will receive will be structured as follows:

**Annotation Data:**

- **id:** Id of the argument in the Debates.org corpus.

- **Argument:** Whole user argument potentially containing biased sentence(s).

- **Sentence:** Extracted sentence of the comment containing the terms of the biased tuple.

- **Biased Tuple:** Biased term tuple found in the provided sentence. It consists of a target term describing a minority demographic group (e.g., "Transgender") and an attribute term describing a stereotyped view of that group (e.g., "pervert").

**Annotation Task:**

- **Biased Sentence:** Annotate as 1 if the sentence is biased, 0 otherwise.

- **Biased Argument:** Annotate as 1 if the whole argument is biased, 0 otherwise.

- **Comment:** If you are unsure how to label the given argument or sentence please describe your problem with the task.

## 3 Example Annotation

**Example 1 (Biased Sentence & Argument):**

Sentence: *"Thus, since being gay is a sin and sins are poor choices, being gay is a choice."* – Annotate as *biased (1)*

Argument: *"If people are born gay, then God would have made them gay. God would have made them to be an abomination. Thus, since being gay is a sin and sins are poor choices, being gay is a choice. Since I have now established that gay acts are a choice and sinful, I can now say that being gay is ungodly. will rest my argument for now."* – Annotate as *biased (1)*

**Example 2 (Unbiased Sentence & Argument):**

Sentence: *"Those who are against homosexual marriage must be immoral as well."* – Annotate as *unbiased (0)*

Argument: *"In fact they were creating a only white race just like they had been taught in evolution. Even today with abortion, Those who say it is wrong must be immoral. Those who are against homosexual marriage must be immoral as well. On both these issues the majority have not had a say in the matter."* – Annotate as *unbiased (0)*

**Example 3 (Unbiased Sentence & Biased Argument):**

Sentence: *"Being gay is not a sin."* – Annotate as *unbiased (0)*

Argument: *"Being gay is not a sin. At all. Rather, acting on those impulses is the sin. Being gay is fine as long as you don't act feelings."* – Annotate as *biased (1)*

**Example 4 (Biased Sentence & Unbiased Argument):**

Sentence: *"Because being gay is a sin."* – Annotate as *biased (1)*

Argument: *"For years, people in the LGBT community have been victim to discrimination, stereotyping, and even bullying and murder. What was the motive? Because being gay is a sin. Because they are unnatural. If it is unnatural for a man to love a man, why is their brain designed to do so?"* – Annotate as *unbiased (0)*

## 4 How to handle confusing cases:

Here are some listings of how to handle potentially confusing cases:

- **Restatement/ Question**: If the user is repeating what his opponent stated or formulates an open question or a question to the opponent. Example: *"Is being gay a sin or a social clique?"* – Annotate as *unbiased (0)*

- **Citations**: If in the extracted sentence, the user is citing a statistic that cannot be checked for validity, please do not provide an annotation.

- **Oversimplified Statements**: If in the extracted sentence, the user is expressing an oversimplified statement without any kind of proof. Example: *"It is common knowledge that gays are the reason for the spread of diseases like HIV."* – Annotate as *biased (1)*

- **Sarcasm**: If the user is sarcastic. Example: *"Imagine if my opponent was the lawyer for the parade: "Your Honor, my client's gay parade is no more repulsive than public executions, the people killed on 9/11, the Sandy Hook mass murder of kids, Rwanda where one million civilians were killed, the Nazi extermination of million Jews.""* – Annotate as *unbiased (0)*

## 5 Additional Notes:

If you face any difficulty labeling the given argument or sentence, please ignore the sub task and ideally provide a description in the comment field as to why this task was difficult to solve.

Please always keep in mind the definition of a stereotype stated in the Introduction of this guide. Argumentation against a minority related topic (e.g., "gay marriage") which does not involve stereotyped statements is not considered a bias.

# Debate.org Islamophobia Annotation Guidelines
# Version 1.0.0

## 1 Introduction

Debate.org is an online debate portal that provides a platform for open discussion, where all members of the community can express their arguments on a wide range of controversial topics. This document describes the annotation guidelines for declaring these user arguments as either expressing stereotyped human bias or not.

According to the Oxford Dictionary of Psychology, a stereotype is defined as:

**"A relatively fixed and oversimplified generalization about a group or class of people."**
(Colman, 2015)

In this annotation task, we explicitly focus on the classification of negative or unfavorable biases.

For this purpose, a CSV file is provided that contains the user arguments, potentially biased sentences from these arguments, and the contained term pairs consisting of a target term referring to a demographic minority group (e.g., "muslim") and an attribute term referring to a presumably biased expression (e.g., "terrorist"). Based on this file, it is up to you to decide whether these extracted sentences or even the entire argument are indeed *biased* or *unbiased*.

If there are open questions that you feel are not covered in this guide, please do not hesitate to get in touch so that these questions can be discussed and the guide improved.

## 2 CSV Presentation

The CSV file you will receive will be structured as follows:

**Annotation Data:**

- **id:** Id of the argument in the Debates.org corpus.

- **Argument:** Whole user argument potentially containing biased sentence(s).

- **Sentence:** Extracted sentence of the comment containing the terms of the biased tuple.

- **Biased Tuple:** Biased term tuple found in the provided sentence. It consists of a target term describing a minority demographic group (e.g., "arabs") and an attribute term describing a stereotyped view of that group (e.g., "oil sheik").

**Annotation Task:**

- **Biased Sentence:** Annotate as 1 if the sentence is biased, 0 otherwise.

- **Biased Argument:** Annotate as 1 if the whole argument is biased, 0 otherwise.

- **Comment:** If you are unsure how to label the given argument or sentence please describe your problem with the task.

## 3 Example Annotation

Example 1 (Biased Sentence & Argument):

Sentence: *"Because this seems to be more about Muslims being violent."* – Annotate as *biased (1)*

Argument: *"Because this seems to be more about Muslims being violent. But there have been refugees from other areas in the world that have settled in fine, like the Jews, and contributed to society. Do you think safe countries should accept refugees if they are not muslim?"* – Annotate as *biased (1)*

**Example 2 (Unbiased Sentence & Argument):**

Sentence: *"I think it is a shame that the right-wing media have indoctrinated people in the west to such an extent that some people automatically equate Islam with terrorism."* – Annotate as *unbiased (0)*

Argument: *"I think it is a shame that the right-wing media have indoctrinated people in the west to such an extent that some people automatically equate Islam with terrorism. There are some Islamic terrorists, true. There are also some catholic terrorists, for example in Northern Ireland. But not all Catholics are terrorists and neither are all Muslims."* – Annotate as *unbiased (0)*

**Example 3 (Unbiased Sentence & Biased Argument):**

Sentence: *"Mostly terrorist attacks committed by Non-Muslims don't happen because of religion, they happen for other reasons."* – Annotate as *unbiased (0)*

Argument: *"Mostly terrorist attacks committed by Non-Muslims don't happen because of religion, they happen for other reasons. Is there people from a certain religion blowing stuff and themselves up as much as Muslims these days? No."* – Annotate as *biased (1)*

**Example 4 (Biased Sentence & Unbiased Argument):**

Sentence: *"It makes sense for people to hate Islam, especially when you consider the terrorist attack of 9/11, and not to mention the Boston bombing, the terrorism in France, etc."* – Annotate as *biased (1)*

Argument: *"I find that there is a distinct difference between being Islamic and being an extremist. It makes sense for people to hate Islam, especially when you consider the terrorist attack of 9/11, and not to mention the Boston bombing, the terrorism in France, etc. I guess what i am trying to say is, aren't we being to harsh on the religion itself?"* – Annotate as *unbiased (0)*

## 4 How to handle confusing cases:

Here are some listings of how to handle potentially confusing cases:

- **Restatement/ Question**: If the user is repeating what his opponent stated or formulates an open question or a question to the opponent. Example: *"Does this mean that every Muslim is going to be a terrorist?"* – Annotate as *unbiased (0)*

- **Citations**: If in the extracted sentence, the user is citing a statistic that cannot be checked for validity, please do not provide an annotation.

- **Oversimplified Statements**: If in the extracted sentence, the user is expressing an oversimplified statement without any kind of proof. Example: *"Islam forces itself with terrorism Islam argues against its own existence.. everyone knows Muslims are terrorists, are you going to for-fit the next round is the only question"* – Annotate as *biased (1)*

- **Sarcasm**: If the user is sarcastic. Example: *"Good luck, trying to to convince me that Islam promotes hate, terrorism, injustice or any other definition of any unpeaceful things."* – Annotate as *unbiased (0)*

## 5 Additional Notes:

If you face any difficulty labeling the given argument or sentence, please ignore the sub task and ideally provide a description in the comment field as to why this task was difficult to solve.

Please always keep in mind the definition of a stereotype stated in the Introduction of this guide. Argumentation against a minority related topic (e.g., "the hijab") which does not involve stereotyped statements is not considered a bias.

